\documentclass[twoside]{article}

\usepackage[accepted]{aistats2020}

\usepackage[utf8]{inputenc} %
\usepackage[T1]{fontenc}    %
\usepackage{hyperref}       %
\usepackage{url}            %
\usepackage{booktabs}       %
\usepackage{amsfonts}       %
\usepackage{nicefrac}       %
\usepackage{microtype}      %
\usepackage{wrapfig}
\usepackage{subcaption}
\usepackage{caption}

\usepackage{float}
\title{Constrained Bayesian Optimization with Max-Value Entropy Search}

\usepackage{amsmath}
\usepackage{bm}
\usepackage{enumerate}
\usepackage{mathtools}
\usepackage{hyperref}
\usepackage{color}
\usepackage{amssymb}
\usepackage{color}
\usepackage{comment}
\usepackage{enumitem}

\usepackage[colorinlistoftodos,prependcaption,textsize=tiny]{todonotes}

\def\xb{{\mathbf x}}

\def\Xcal{\mathcal{X}}

\newcommand{\field}[1]{\mathbb{#1}}

\newcommand{\R}{\field{R}}

\newcommand{\vect}[1]{\boldsymbol{#1}} 
\newcommand{\mat}[1]{\boldsymbol{#1}} 
\newcommand{\tvect}[1]{\tilde{\boldsymbol{#1}}}
\newcommand{\tmat}[1]{\tilde{\boldsymbol{#1}}}
\newcommand{\tscal}[1]{\tilde{#1}}
\newcommand{\hvect}[1]{\hat{\boldsymbol{#1}}}
\newcommand{\hmat}[1]{\hat{\boldsymbol{#1}}}
\newcommand{\hscal}[1]{\hat{#1}}
\newcommand{\bvect}[1]{\bar{\boldsymbol{#1}}}
\newcommand{\bmat}[1]{\bar{\boldsymbol{#1}}}
\newcommand{\bscal}[1]{\bar{#1}}
\newcommand{\vzero}{\vect{0}}
\newcommand{\vone}{\vect{1}}

\newcommand{\dummystring}{QWERTYU}
\newcommand{\vci}[3][\dummystr]{\ifthenelse{\equal{#1}{\dummystring}}{\vect{#2}_{#3}}{\vect{#2}_{#3}^{(#1)}}}
\newcommand{\mx}[3][\dummystr]{\ifthenelse{\equal{#1}{\dummystring}}{\mat{#2}_{#3}}{\mat{#2}_{#3}^{(#1)}}}
\newcommand{\tvci}[3][\dummystr]{\ifthenelse{\equal{#1}{\dummystring}}{\tvect{#2}_{#3}}{\tvect{#2}_{#3}^{(#1)}}}
\newcommand{\tmx}[3][\dummystr]{\ifthenelse{\equal{#1}{\dummystring}}{\tmat{#2}_{#3}}{\tmat{#2}_{#3}^{(#1)}}}
\newcommand{\tscl}[3][\dummystr]{\ifthenelse{\equal{#1}{\dummystring}}{\tscal{#2}_{#3}}{\tscal{#2}_{#3}^{(#1)}}}
\newcommand{\hvci}[3][\dummystr]{\ifthenelse{\equal{#1}{\dummystring}}{\hvect{#2}_{#3}}{\hvect{#2}_{#3}^{(#1)}}}
\newcommand{\hmx}[3][\dummystr]{\ifthenelse{\equal{#1}{\dummystring}}{\hmat{#2}_{#3}}{\hmat{#2}_{#3}^{(#1)}}}
\newcommand{\hscl}[3][\dummystr]{\ifthenelse{\equal{#1}{\dummystring}}{\hscal{#2}_{#3}}{\hscal{#2}_{#3}^{(#1)}}}
\newcommand{\bvci}[3][\dummystr]{\ifthenelse{\equal{#1}{\dummystring}}{\bvect{#2}_{#3}}{\bvect{#2}_{#3}^{(#1)}}}
\newcommand{\bmx}[3][\dummystr]{\ifthenelse{\equal{#1}{\dummystring}}{\bmat{#2}_{#3}}{\bmat{#2}_{#3}^{(#1)}}}
\newcommand{\bscl}[3][\dummystr]{\ifthenelse{\equal{#1}{\dummystring}}{\bscal{#2}_{#3}}{\bscal{#2}_{#3}^{(#1)}}}

\newcommand{\Ex}{\mathrm{E}}
\newcommand{\Var}{\mathrm{Var}}

\newcommand{\Ind}[1]{\mathrm{I}_{\{#1\}}}

\newcommand{\Ent}{\mathrm{H}}

\newcommand{\Id}{\mat{I}}

\newcommand{\figref}[1]{Figure~\ref{fig:#1}}
\newcommand{\tabref}[1]{Table~\ref{tab:#1}}
\newcommand{\secref}[1]{Section~\ref{sec:#1}}
\renewcommand{\eqref}[1]{Eq.~\ref{eq:#1}}
\newcommand{\eqp}[1]{(\ref{eq:#1})}

\newcommand{\tsgamma}[2][\dummystring]{\tscl[#1]{\gamma}{#2}}

\newcommand{\tskappa}[2][\dummystring]{\tscl[#1]{\kappa}{#2}}

\newcommand{\vk}[2][\dummystring]{\vci[#1]{k}{#2}}

\newcommand{\vp}[2][\dummystring]{\vci[#1]{p}{#2}}

\newcommand{\vx}[2][\dummystring]{\vci[#1]{x}{#2}}

\newcommand{\vz}[2][\dummystring]{\vci[#1]{z}{#2}}

\newcommand{\hvmu}[2][\dummystring]{\hvci[#1]{\mu}{#2}}

\newcommand{\mxk}[2][\dummystring]{\mx[#1]{K}{#2}}
\newcommand{\mxl}[2][\dummystring]{\mx[#1]{L}{#2}}
\newcommand{\mxm}[2][\dummystring]{\mx[#1]{M}{#2}}
\newcommand{\mxn}[2][\dummystring]{\mx[#1]{N}{#2}}

\newcommand{\mxx}[2][\dummystring]{\mx[#1]{X}{#2}}

\newcommand{\hmxl}[2][\dummystring]{\hmx[#1]{L}{#2}}

\newcommand{\hmxy}[2][\dummystring]{\hmx[#1]{Y}{#2}}

\newcommand{\hmxsigma}[2][\dummystring]{\hmx[#1]{\Sigma}{#2}}

\begin{document}

\twocolumn[

\aistatstitle{Constrained Bayesian Optimization with Max-Value Entropy Search}

 \aistatsauthor{ Valerio Perrone, Iaroslav Shcherbatyi, Rodolphe Jenatton\footnotemark  \addtocounter{footnote}{-1}, C\'{e}dric Archambeau, Matthias Seeger}
 \aistatsaddress{Amazon \\ Berlin, Germany \\   \texttt{ \{vperrone, siarosla, cedrica, matthis\}@amazon.com}}

 ]
 
 \footnotetext{$^*$Work done while affiliated with Amazon; now at Google Brain, Berlin, \texttt{rjenatton@google.com}.}

\begin{abstract}
Bayesian optimization (BO) is a model-based approach to sequentially optimize expensive black-box functions, such as the validation error of a deep neural network with respect to its hyperparameters. In many real-world scenarios, the optimization is further subject to a priori unknown constraints. For example, training a deep network configuration may fail with an out-of-memory error when the model is too large. In this work, we focus on a general formulation of Gaussian process-based BO with continuous or binary constraints. We propose constrained Max-value Entropy Search (cMES), a novel information theoretic-based acquisition function implementing this formulation. We also revisit the validity of the factorized approximation adopted for rapid computation of the MES acquisition function, showing empirically that this leads to inaccurate results. On an extensive set of real-world constrained hyperparameter optimization problems we show that cMES compares favourably to prior work, while being simpler to implement and faster than other constrained extensions of Entropy Search.
\end{abstract}

\section{Introduction}
\label{sec:intro}

Consider the problem of tuning the hyperparameters of a large neural network to minimize its validation error. This validation error is a {\em black-box} in that neither its analytical form nor gradients are available, and each (noisy) point evaluation requires time-consuming training from scratch. In Bayesian optimization (BO) the black-box function $y(\xb)$ is queried sequentially at points selected by optimizing an acquisition function, based on a probabilistic surrogate model (e.g., a Gaussian process) fit to the evaluations collected so far \cite{Mockus1978, Jones1998, Shahriari2016}.
In many real-world settings, this black-box optimization problem is subject to stochastic constraints. For example, we may want to maximize the accuracy of a machine learning model while limiting its training time or prediction latency. Here, objective and constraint are {\em real-valued} functions which are jointly observed. Alternatively, we may want to tune a deep neural network (DNN) while avoiding training failures due to out-of-memory (OOM) errors. In this latter case, the constraint feedback is {\em binary}, and the objective is not observed at points where the constraint is violated.

It is custom to treat a constraint on par with the objective, using a joint or conditionally independent random function model. On the one hand, unfeasible evaluations carry a cost (e.g., wasted resources, compute node failure), so their occurence should be minimized. On the other hand, avoiding the unfeasible region should not lead to convergence to suboptimal hyperparameters. Often, good configurations lie at the boundary of the feasible region. For example, setting a high learning rate when training a DNN can lead to better performance but could make the training diverge. Ideally, a user should be able to control the probability of the final solution satisfying the constraints.

In the absence of constraints, Gaussian process (GP) based BO typically uses simple acquisition functions, such as Expected Improvement (EI) \cite{Mockus1978} or Upper Confidence Bound (UCB) \cite{Srinivas:12}. Unfortunately, their extension to the constrained case poses conceptual difficulties \cite{Gardner14, Gelbart14, Lobato15a}. Entropy Search (ES) \cite{Hennig:12} or Predictive Entropy Search (PES) \cite{Hernandez:16, Lobato2016} render acquisition functions tailored to constrained BO \cite{Lobato15a}. However, these are much more complex and expensive to evaluate than EI. Max-value Entropy Search (MES) was recently introduced as a simple and efficient alternative to PES \cite{Wang:17}, but has not been extended to handle stochastic constraints.

In this work, we focus on BO with continuous or binary feedback constraints. The continuous feedback case is simpler to implement and has received much attention in the literature, while binary feedback constraints are underserved in prior work despite their relevance in practice. We develop Max-value Entropy Search with constraints (cMES), a novel information theoretic acquisition function that generalizes MES. cMES can handle both continuous and binary feedback, while retaining the computational efficiency of its unconstrained counterpart \cite{Wang:17}. %
We evaluate cMES on a range of constrained black-box optimization problems, demonstrating that it tends to outperform previously published methods
such as constrained EI~\cite{Gardner14}. We also consider a simple, but strong {\em adaptive percentile} baseline, which is an improved variant of the high-value heuristic from \cite{Gelbart14}. Despite its simplicity, it can outperform constrained EI. We therefore encourage the community to also consider this new baseline in future research.

As a byproduct of our evaluations, we analyze the impact of different ways of sampling the (constrained) maximum $y_{\star}$ from the posterior distribution. We find that the independent ``mean field'' approximation proposed in \cite{Wang:17} leads to poor results, providing an explanation for this finding. Jointly dependent sampling works well and remains tractable in our experiments, even though it scales cubically in the size of the discretization set.

The remainder of the paper is organized as follows. We review the background and related work in Section~\ref{sec:related work}, introduce the proposed methodology in Section~\ref{sec:mes-constraints}, present experimental results in Section~\ref{sec:experiments}, and outline conclusions and further developments in Section~\ref{sec:conclusion}.

\section{Constrained Bayesian Optimization}
\label{sec:related work}

Let $y(\xb): \Xcal \rightarrow \mathbb{R}$ represent a black-box function over a set $\Xcal \subset \mathbb{R}^p$. For instance, $y(\xb)$ is the validation error of a deep neural network (DNN) as a function of its hyperparameters $\xb$ (e.g., learning rate, number of layers, dropout rates). Each evaluation of $y(\xb)$ requires training the network, which can be expensive to do. Our aim is to minimize $y(\xb)$ with as few queries as possible. Bayesian optimization (BO) is an efficient approach to find a minimum of the black-box function $y(\xb)$, where $\xb \in \mathcal{X}$ \cite{Jones1998,Eggensperger2012,Shahriari2016}. The idea is to replace $y(\xb)$ by a Gaussian process surrogate model \cite{Rasmussen2006, Shahriari2016}, updating this model sequentially by querying the black-box at new points. Query points are found by optimizing an acquisition function, which trades off exploration and exploitation. However, conventional models and acquisition functions are not designed to take constraints into account.

Our goal is to minimize the target black-box $y(\xb)$, subject to a constraint $c(\vx{}) \le \delta$. In this paper, we limit our attention to modeling the feasible region by a single function $c(\xb)$. As the latent constraints are pairwise conditionally independent, an extension to multiple constraints is straightforward, yet notationally cumbersome. Both $y(\xb):\Xcal \rightarrow \mathbb{R}$ and $c(\xb): \Xcal \rightarrow \mathbb{R}$, are unknown and need to be queried sequentially. The constrained optimization problem we would like to solve is defined as follows:
\begin{equation}\label{eq:ystar}
   y_{\star} = \min_{\vx{}\in\mathcal{X}}\left\{ y(\vx{})\; \|\; c(\vx{}) \le \delta \right\} ,
\end{equation}
where $\delta\in\mathbb{R}$ is a confidence parameter. The latent functions $y(\xb)$ and $c(\xb)$ are assumed to be conditionally independent in our surrogate model, with different GP priors placed on them.

We consider two different setups, depending on what information is observed about the constraint. Most previous work \cite{Gardner14, Gelbart14, Lobato15a} assumes that {\em real-valued feedback} is obtained on $c(\xb)$, just as for $y(\xb)$. In this case, both latent function can be represented as GPs with Gaussian noise. Unfortunately, this setup does not cover practically important use cases of constrained hyperparameter optimization. For example, if training a DNN fails with an out-of-memory (OOM) error, we cannot observe the amount of memory requested just before the crash, neither do we usually know the exact amount of memory available on the compute instance in order to calibrate $\delta$. Covering such use cases requires handling {\em binary feedback} on $c(\xb)$, even though this is technically more difficult. An evaluation returns $z_y\sim N(z_y | y(\vx{}), \alpha_y^{-1})$ and $z_c\in\{-1, +1\}$, where $z_c = -1$ for a feasible, $z_c = +1$ for an unfeasible point. We never observe the latent constraint function $c(\xb)$ directly. We assume $z_c\sim \sigma(z_c c(\vx{}))$, where $\sigma(t) = \frac{1}{1 + e^{-t}}$ is the logistic sigmoid, but other choices are possible. We can then rewrite the constrained optimization problem (\ref{eq:ystar}) as follows:
\[
   y_{\star} = \min_{\vx{}\in\mathcal{X}} \left\{ y(\vx{})\; \|\; P(z_c = +1 | \vx{}) =
  \sigma(c(\vx{})) \le \sigma(\delta) \right\}.
\]
This formulation is similar to the one proposed in \cite{Gelbart14}. The confidence parameter $\sigma(\delta)\in (0,1)$ controls the size of the (random) feasible region for defining $y_{\star}$. Finally, note that in the example of OOM training failures, the criterion observation $z_y$ is obtained only if $z_c = -1$: if a training run crashes, a validation error is not obtained for the queried configuration. Apart from a single experiment in \cite{Gelbart14}, we are not aware of previous contrained BO work covering the binary feedback case, as we do here.

\subsection*{Related Work}

The most established technique to tackle constrained BO is constrained EI (cEI) \cite{Gardner14, Gelbart14, Snoek2015, letham2019}. If the constraint is denoted by $c(\xb)\le 0$, a separate regression model is used to learn the constraint function $c(\xb)$ (typically a GP), and EI is modified in two ways. First, the expected amount of improvement of an evaluation is computed only with respect to the current \emph{feasible} minimum. Second, hyperparameters with a large probability of satisfying the constraint are encouraged by optimizing $cEI(\xb) = P\{ c(\xb)\le 0\} EI(\xb)$, where $P\{ c(\xb)\le 0\}$ is the posterior probability of $\xb$ being feasible under the constraint model, and $EI(\xb)$ is the standard EI acquisition function. %

Several issues with cEI are detailed in \cite{Lobato15a}. First, the current feasible minimum has to be known, which is problematic if all initial evaluations are unfeasible. A workaround is to use a different acquisition function initially, focussed on finding a feasible point \cite{Gelbart14}. In addition, the probability of constraint violation is not explicitely accounted for in cEI. A confidence parameter equivalent to ours features in \cite{Gelbart14}, but is only used to recommend the final hyperparameter configuration. %
Another approach was proposed in \cite{Lobato15a}, where PES is extended to the constrained case. Constrained PES (cPES) can outperform cEI and does not require the workarounds mentioned above. However, it is complex to implement, expensive to evaluate, and unsuitable for binary constraint feedback. %

Drawing from numerical optimization, different generalizations of EI to the constrained case are developed in \cite{Picheny2016} and \cite{Ariafar2019}. The authors represent the constrained minimum by way of Lagrange multipliers, and the resulting query selection problem is solved as a sequence of unconstrained problems. These approaches are not designed for the binary constraint feedback, and either require numerical quadrature~\cite{Picheny2016} or come at the cost of a large set of extra hyperparameters~\cite{Ariafar2019}. In contrast, our acquisition function can be optimized by a standard unconstrained optimizer, thus is simple to integrate into BO packages such as GPyOpt~\cite{Gpyopt2016} .

\section{Max-Value Entropy Search with Constraints}
\label{sec:mes-constraints}

In this section we derive cMES, a novel max-value entropy search acquisition function scoring the value of an evaluation at some $\xb\in\Xcal$. Our method supports, both, real-valued and binary constrained feedback. Since the binary feedback case is more challenging to derive, and is not covered by prior work, we here focus on $z_c\in\{-1, +1\}$, where $P(z_c = +1 | \xb) = \sigma(c(\xb))$. The derivation for $z_c\in\R$ is provided in Section~\ref{sec:real-valued-feedback} of the supplemental material.

We assume $y(\cdot)$ and $c(\cdot)$ are given independent Gaussian process priors, with mean zero and covariance functions $k_y(\vx{}, \vx{}')$ and $k_c(\vx{}, \vx{}')$ respectively. Moreover, data $\mathcal{D}=\{ \xb_i, z_{y i}, z_{c i} \| i=1,\dots, n \}$ has already been acquired. Since $z_{y i}\sim N(y(\xb_i), \alpha_y^{-1})$, the posterior for $y(\cdot)$ is a GP again \cite{Rasmussen2006}, with marginal mean and variance given by
\begin{align*}
  \mu_y(\vx{}) &= \vk{y}(\vx{})^T \mxm{}^{-1}\vz{y}, \\
  \sigma_y^2(\vx{}) &=
  k_y(\vx{}, \vx{}) - \vk{y}(\vx{})^T \mxm{}^{-1} \vk{y}(\vx{}),
\end{align*}
where $\vz{y} = [ z_{y i} ]\in\R^n$, $\mxm{} = [k_y(\vx{i},\vx{j})] + \alpha_y^{-1}\Id\in\R^{n\times n}$, and $\vk{y}(\vx{}) = [k_y(\vx{}, \vx{i})]\in\R^n$. For real-valued constraint feedback (i.e., $z_{c i}\in\R$), we can use the same formalism for the posterior over $c(\cdot)$. In the binary feedback case, we use expectation propagation \cite{Minka:01a} in order to approximate the posterior for $c(\cdot)$ by a GP. In the sequel, we denote the posterior marginals of these processes at input $\xb$ by $P(y) = N(y | \mu_y, \sigma_y^2)$ and $P(c) = N(c | \mu_c, \sigma_c^2)$, dropping the conditioning on $\mathcal{D}$ and $\xb$ for convenience. Details on $\mu_c(\vx{})$, $\sigma_c^2(\vx{})$ are given in \cite[Sect.~3.6.1]{Rasmussen2006}.

The unconstrained MES acquisition function is given by
\begin{align*}
  \mathcal{I}( y ; y_{\star} ) = \Ent[ P(y) ] - \Ex_{y_{\star}}\left[ \Ent[ P(y | y_{\star}) ] \right],
\end{align*} 
where the expectation is over $P(y_{\star} | \mathcal{D})$, and $y_{\star} = \min_{\vx{}\in\mathcal{X}} y(\vx{})$ \cite{Wang:17}. Here, $\Ent[ P(y) ] = \int P(y) (-\log P(y))\, d y$ denotes the differentiable entropy and $P(y | y_{\star}) \propto P(y) \Ind{y\ge y_{\star}}$ is a truncated Gaussian. First, it should be noted that this is a simplifying assumption. In PES \cite{Hernandez:16}, the related distribution $P(y | \vx{\star})$ is approximated, where $\vx{\star}$ is the argmin. Several local constraints on $y(\cdot)$ at $\vx{\star}$ are taken into account, such as $\nabla_{\vx{\star}} y = \vzero$. This is not done in MES, which simplifies derivations considerably. Second, the expectation over $y_{\star}$ is approximated by Monte Carlo sampling.

The cMES acquisition we develop is a generalization of MES. For binary feedback, this extension modifies the mutual information criterion as follows:
\[
  \mathcal{I}( (y, z_c) ; y_{\star} ) = \Ent[ P(y, z_c) ] - \Ex_{y_{\star}}\left[ \Ent[ P(y, z_c | y_{\star}) ]
  \right],
\]
where $y_{\star}$ is the constrained minimum from (\ref{eq:ystar}). Note that we use the noise-free $y$ in place of $z_y$ for simplicity, as done in \cite{Wang:17}. A variant incorporating Gaussian noise is described in Section~\ref{sec:gaussian-noise} of the supplemental material.
We first show how to approximate the entropy difference for fixed $y_{\star}$, then how to sample from $P(y_{\star} | \mathcal{D})$ in order to approximate $\Ex_{y_{\star}}[\cdot]$ by Monte Carlo.

\subsection*{Entropy Difference for fixed $y_{\star}$}

Suppose that we observed $c$ in place of $z_c$. What would $P(y, c | y_{\star})$ be? If $c\le \delta$, then $y\ge y_{\star}$, while if $c > \delta$, our belief in $y$ remains the same. In other words, $P(y, c | y_{\star}) = Z^{-1} P(y) P(c) \kappa(y, c)$, where $\kappa(y, c) = 1 - \Ind{c\le\delta} \Ind{y\le y_{\star}}$ is an indicator function. The entropy difference $\Ent[ P(y,  c) ] - \Ent[ P(y, c | y_{\star}) ]$ can now be expressed in terms of
\begin{align*}
  \gamma_c &= \frac{\delta - \mu_c}{\sigma_c},  \quad
   \gamma_y = \frac{y_{\star} -
  \mu_y}{\sigma_y}, \\
  Z_c &= \Ex[\Ind{c\le\delta}] = \Phi(\gamma_c),  \quad
  Z_y = \Ex[\Ind{y\le y_{\star}}] = \Phi(\gamma_y) ,
\end{align*}
where $\Phi(t) = \Ex[\Ind{n\le t}]$ and $n\sim N(0,1)$ is the cumulative distribution function for a standard normal variate. For example, $Z = \Ex[\kappa(y, c)] = 1 - Z_c Z_y$. Details are given in Section~\ref{sec:real-valued-feedback} of the supplemental material, essentially providing the cMES derivation for real-valued constraint feedback.

For a binary response $z_c\in\{\pm 1\}$, we need to take into account that less information about $y_{\star}$ is obtained. Since $P(z_c | c) = \sigma(z_c c)$ is not Gaussian, we approximate
\begin{align*}
  Q(z_c) Q(c | z_c) & \approx P(z_c | c) P(c), \quad z_c  \in\{\pm 1\},
\end{align*}
where the $Q(c | z_c)$ are Gaussians. We make use of Laplace's approximation, in particular the accurate approximation $Q(z_c)\approx P(z_c)$ is detailed in \cite[Sect.~4.5.2]{Bishop:06}. Now:
\begin{align*}
\begin{split}
   P(y, z_c | y_{\star})  & = \int P(y) P(z_c | c) P(c) \kappa(y, c)\, d c \\
    &\approx \int P(y) Q(z_c) Q(c | z_c) \kappa(y, c)\, d c \\
   & =  P(y) Q(z_c) \tilde{\kappa}(y, z_c),\; \\ 
   \tilde{\kappa}(y, z_c)  & = 1 - \Ind{y\le y_{\star}} F(z_c),\; \\
   F(z_c) & = \Ex_{Q(c | z_c)}[\Ind{c\le \delta}].
\end{split}
\end{align*}
While $\tilde{\kappa}(y, z_c)$ is not an indicator, it is piece-wise constant, allowing for an analytically tractable computation of the entropy difference:
\begin{align*}
  & \Ent[P(y)] + \Ent[Q(z_c)] - \Ent[P(y, z_c | y_{\star})] = \\
 & -\log Z - B\left( \gamma_y h(-\gamma_y)/2 + \tilde{Z}_c^{-1} \Ex_Q\left[ (1 - F(z_c))  \right. \right.  \\
 & \left.  \left.  (-\log(1 - F(z_c))) +
  (F(z_c) - \tilde{Z}_c) \log Q(z_c) \right] \right), \\
  & B = Z_y \tilde{Z}_c Z^{-1} = \left( \exp(-\log Z_y -\log \tilde{Z}_c) - 1 \right)^{-1} ,
\end{align*}
where $F(z_c) = \Ex_{Q(c | z_c)}[\Ind{c\le \delta}]$, $\tilde{Z}_c = \Ex_Q[F(z_c)]$, and $Z = 1 - Z_y \tilde{Z}_c$. Function $h(x) = N(x | 0,1) / \Phi(-x)$ denotes the hazard function for the standard normal distribution. A derivation of this expression is given in Section~\ref{sec:entdiff-binary} of our supplemental material, along with recommendations for a numerically robust implementation. All terms depending on $c$ and $z_c$ are independent of $y_{\star}$, and can therefore be precomputed.

\subsection*{Sampling from $P(y_{\star} | \mathcal{D})$}

In the constrained case, we aim to sample from $P(y_{\star} | \mathcal{D})$, where $y_{\star} = \min_{\vx{}\in\mathcal{X}}\left\{ y(\vx{})\; \|\; c(\vx{}) \le \delta \right\}$. Here, $y(\cdot)$ and $c(\cdot)$ are posterior GPs conditioned on the current data $\mathcal{D}$. This primitive is known as {\em Thompson sampling} for a GP model \cite{Thompson:33, Kandasamy:18}. For commonly used infinite-dimensional kernels, it is intractable to draw exact sample functions from these GPs, let alone to solve the conditional optimization problem for $y_{\star}$. Several approximations have been considered in prior work.

In \cite{Lobato15a}, a finite-dimensional random kitchen sink (RKS) approximation is used to draw approximate sample paths, and the constrained problem is solved for these. Since the RKS basis functions are nonlinear in $\vx{}$, so are the objective and constraint function, and solving for $y_{\star}$ requires complex machinery. Moreover, each kernel function has a different RKS expansion, and the latter is not readily available for many kernels used in practice. A simpler approach is used in \cite{Wang:17}. They target the cumulative distribution function (CDF) of $y_{\star}$, which can be written as expectation over $y(\cdot)$ and $c(\cdot)$ of an infinite product. This is approximated by restricting the product over a finite set $\hat{\mathcal{X}}$, and by assuming {\em independence} of all $y(\vx{})$ and $c(\vx{})$ for $\vx{}\in \hat{\mathcal{X}}$. Given these assumption, their CDF approximation can be evaluated by computing marginal posteriors over points in $\hat{\mathcal{X}}$, which scales linearly in $|\hat{\mathcal{X}}|$.

\begin{figure}[ht]
\begin{center}
\includegraphics[width=0.6\columnwidth]{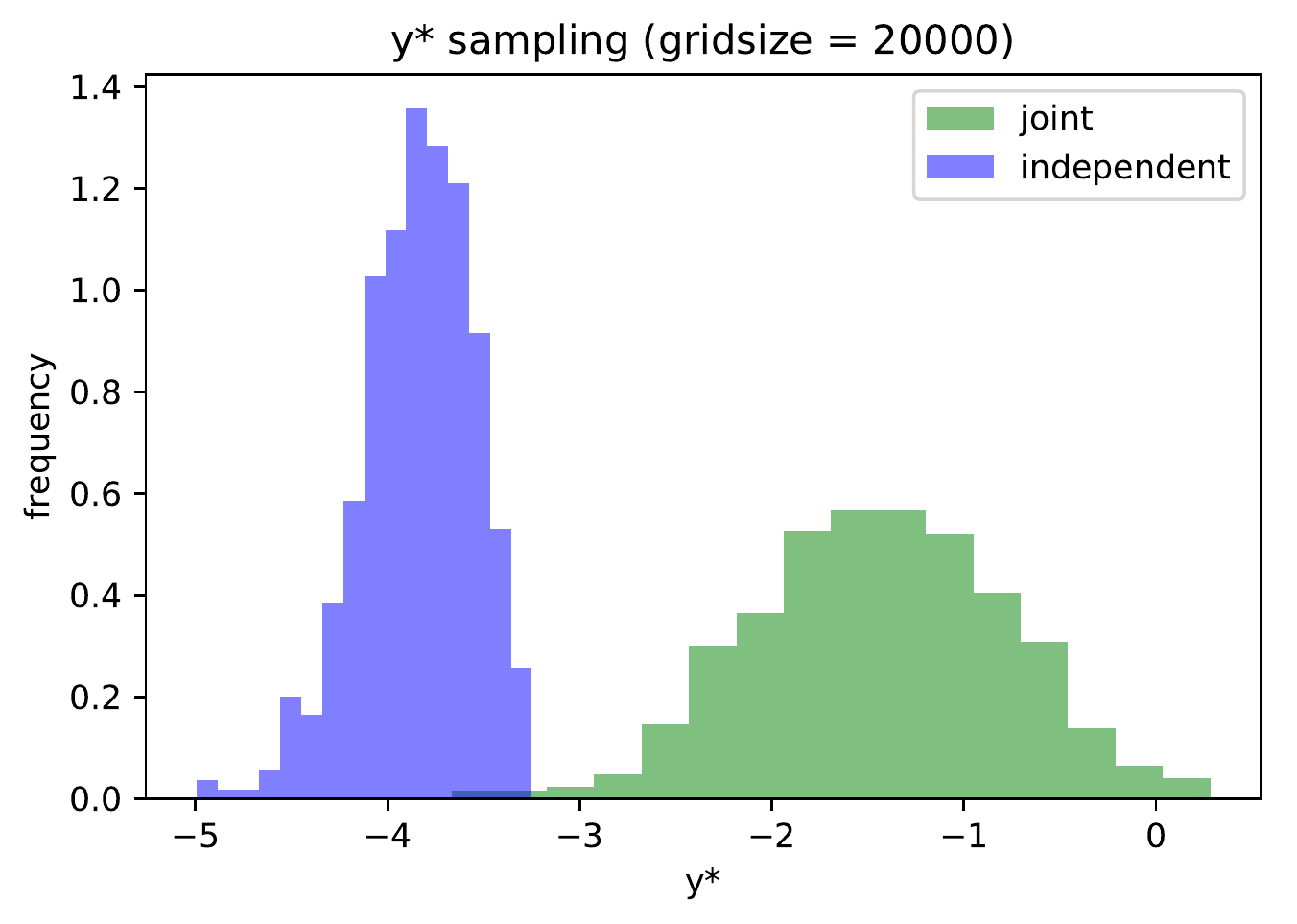}
\caption{\label{fig:ystar_histogram}
  Distribution of the sampled blackbox minimum $y_{\star}$ obtained via the ground truth (joint sampling) in contrast with the highly biased samples obtained via independent sampling. In the independent sampling case, the sampled objective minima do not become more accurate as more points in $\hat{\mathcal{X}}$ are used, but diverge to -$\infty$.}
\end{center}
\end{figure}

While this gives rise to a tractable approximation of the CDF, we found this approximation to result in poor overall performance in our experiments. A simple alternative is to restrict ourselves to a finite set $\hat{\mathcal{X}}$ (we use a Sobol sequence \cite{Sobol1967}), but then draw {\em jointly dependent} samples of $y(\hat{\mathcal{X}})$ and $c(\hat{\mathcal{X}})$ respectively, based on which $y_{\star}$ (restricted to $\hat{\mathcal{X}}$) is trivial to compute (Section~\ref{sec:sampling-ystar} in the supplemental material). While joint sampling scales cubically in the size of $\hat{\mathcal{X}}$, sampling takes less than a second for $|\hat{\mathcal{X}}| = 2000$, the size we used in our experiments. 

In Table~\ref{tab:marginal}, we compare BO for different variants of cMES, using joint or marginal sampling of $y_{\star}$ respectively. It is clear that joint sampling leads to significantly better results across the board. As noted in \cite{Wang:17}, $y_{\star}$ drawn under their independence assumption is underbiased. We show this in Figure~\ref{fig:ystar_histogram}, where the size of this bias is very significant. Importantly, the bias gets {\em worse} the larger $\hat{\mathcal{X}}$ is: $y_{\star}$ diverges as $|\hat{\mathcal{X}}|\to \infty$. This means that the regime of $|\hat{\mathcal{X}}|$ where the bias is small enough not to distort results, is likely small enough to render joint sampling perfectly tractable. While for high-dimensional configuration spaces, a discretization set of size $|\hat{\mathcal{X}}| = 2000$ may prove insufficient, and more complex RKS approximations may have to be used \cite{Lobato15a}, the simple jointly dependent Thompson sampling solution should always be considered as a baseline.\footnote{
   Jointly dependent sampling is also used in \cite{Kandasamy:18} (personal communication).}

\section{Experiments}
\label{sec:experiments}

In this section, we compare our novel cMES acquisition function against competing approaches in a variety of settings. We start with {\em binary feedback} scenarios, where the latent constraint function $c(\cdot)$ is accessed indirectly via $z_c\in\{-1, +1\}$, the model for $c(\cdot)$ uses a Bernoulli likelihood, and inference is approximated by expectation propagation \cite{Minka:01a, Rasmussen2006}. We consider two variants: {\em observed-objective}, where the objective $y(\cdot)$ is observed with each evaluation (feasible or not); and {\em unobserved-objective}, where an observation $z_y$ is obtained only if $z_c = -1$ (feasible). We also work on {\em real-valued feedback} scenarios, where $c(\cdot)$ is observed directly via $z_c\in\R$, and inference for the $c(\cdot)$ model is analytically tractable. In this latter case, we compare against a larger range of prior work, extensions of which to the binary feedback case are not available.

In all scenarios, we compare against constrained EI (cEI) \cite{Gelbart14}, which can be used with binary feedback. Since EI needs a feasible incumbent, we minimize the probability of being unfeasible as long as no feasible points have been observed \cite{Gelbart14}. As baselines, we use random search \cite{Bergstra2012}, as well as a novel heuristic called {\em adaptive percentile} (AP). AP is a variant of the high-value heuristic introduced in \cite{Gelbart14}, where a single GP $y(\cdot)$ is used. Whenever an evaluation is unfeasible, AP plugs in the $p$-percentile of all previously observed objective values as target value. We consider $perc\ge 50$, noting that $perc=100$ corresponds to plugging in the maximum observed so far. We compare against PESC~\cite{Lobato15a} for real-valued feedback, as it does not support binary feedback.

In all experiments, we compute cMES by drawing the constrained optimum $y_{\star}$ via jointly dependent Thompson sampling as detailed in \secref{mes-constraints}, using a discretization set  $\hat{\Xcal}$ of size 2000. The subscript $observe$ indicates the observed-objective scenario ($z_y$ always observed), and for cMES $p$ denotes $\sigma(\delta)$. Unless said otherwise, methods are implemented in GPyOpt~\cite{Gpyopt2016}, using a Mat\'ern $\frac{5}2$ covariance kernel with automatic relevance determination hyperparameters, optimized by empirical Bayes \cite{Rasmussen2006}. Integer-valued hyperparameters are dealt with by rounding to the closest integer after continuous optimization of the acquisition function, and one-hot encoding is used for categorical variables.

To build up intuition, we start with an artificial 2D constrained optimization problem. We then compare all methods on a range of ten constrained hyperparameter optimization (HPO) problems, involving real-world datasets and machine learning methods from {\tt scikit-learn} \cite{pedregosa2011scikit}.

\subsection{2D Constrained Optimization Problem}

We compared the behavior of cMES and cEI on a 2D constraint optimization problem. The synthetic black-box function consists of three quadratics in 2D ($\Xcal = [-1, 1]^2$) with a disconnected feasible region, the smallest component of which contains the global optimum. Formally, $y(x, y) = \min \left( {\scriptstyle \frac{(x_1 - x)^2 + (y_1 - y)^2}{0.02} +  0.3,  \frac{(x_2 - x)^2 + (y_2 - y)^2}{0.2}  +  0.6 },  \right. \\
 \left. {\scriptstyle \frac{(x_3 - x)^2 + (y_3 - y)^2}{0.6} + 0.9  } \right)$, with $(x_1, y_1) = (-0.7, 0.5), (x_2, y_2) = (0.5, 0.3),$ and $(x_3, y_3) = (-0.3, -0.3)$. We consider the unobserved-objective scenario, and define the feasible region by $y(x, y) < 1.2$. We warm-started both cEI and cMES with the same 5 points randomly drawn from the search space, and we ran constrained BO under the same fixed budget.

\begin{figure*}[t]
\begin{center}
\begin{subfigure}[b]{1\textwidth}
   \includegraphics[width=1\linewidth]{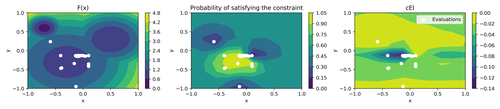}
   \caption{}
   \label{fig1} 
\end{subfigure}
\begin{subfigure}[b]{1\textwidth}
   \includegraphics[width=1\linewidth]{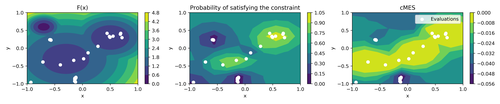}
   \caption{}
   \label{fig2} 
\end{subfigure}
\caption{2D toy example. Visualization of the evaluations proposed by cEI (above) and cMES (below). Both methods were initialized with the same 5 random points. The areas with objective values greater than 1.2 are unfeasible, and centers of the three dark valleys correspond to two local and one global optima.}
\label{figure-2d-example}
\end{center}
\end{figure*}

Figure~\ref{figure-2d-example} shows the objective surface, the probability of satisfying the constraint, and the acquisition function values, for both the cEI and the cMES. As the objective value is unobserved upon violations of the constraint, the GP model placed on the objective is not updated after querying unfeasible points. The cMES is able to better account for the constraint surface and finds the valley in the upper right corner, converging to a better solution.

\subsection{Ten real-world hyperparameter tuning problems}

We then considered ten constrained HPO problems, spanning different {\tt scikit-learn} algorithms \cite{pedregosa2011scikit}, {\tt libsvm} datasets \cite{Chang2011}, and constraint modalities. The first six problems are about optimizing an accuracy metric (AUC for binary classification, coefficient of determination for regression), subject to a constraint on model size, a setup motivated by applications in IOT or on mobile devices. The remaining four problems require minimizing the error on positives, subject to a limit on the error on negatives as is relevant in medical domain applications. A summary of algorithms, datasets, and fraction of feasible configurations is given in \tabref{blackbox-description}. In the table, we denote as $d$ the input dimension description of the blackbox function, in the following format: total number of dimensions (number of real dimensions, number of integer valued dimensions, number of categorical dimensions). 

When sampling a problem, and then a hyperparameter configuration at random, we hit a feasible point with probability 51.5\%. Also note that for all these problems, the overall global minimum point is unfeasible. Further details about these problems and the choice of thresholds for the constraints are provided in Section~\ref{sec:experiment-details} of the supplement.

We ran each method to be compared on the ten HPO problems described above, using twenty random repetitions each. We start each method with evaluations at five randomly sampled candidates. To account for the heterogeneous scales of the 10 blackboxes and be able to compare the relative performance of the competing methods, common practice is to aggregate results based on the {\em average rank} (lower, better) \cite{Feurer:15, Bardenet2013, klein19}. Specifically, we rank methods for the same HPO problem, iteration, and random seed according to the best feasible value they observed so far, then average over all these. Note that in initial rounds, some methods may not have made feasible observations. For example, if five of ten methods have feasible evaluations, then the former are ranked $1,\dots, 5$, while the latter are equally ranked $(6+10)/2 = 8$. 

\begin{figure}[H]
\includegraphics[width=0.46\textwidth]{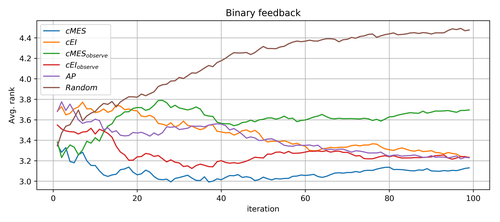}
\caption{\label{fig:results_ranking}
   Binary feedback. Average rank per iteration for the best-performing methods in each
   category.}
\end{figure}

\begin{figure}[H]
\includegraphics[width=0.46\textwidth]{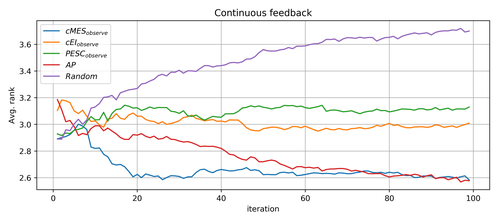}
\caption{\label{fig:results_ranking_cont}
   Real-valued feedback. Average rank per iteration for the best-performing methods in each
   category.}
\end{figure}

The results for the binary-feedback case in \tabref{results} and \figref{results_ranking} point to a number of conclusions. First, among methods operating in the unobserved scenario, cMES achieves the best overall average rank. While cEI uses fewer unfeasible evaluations, it is overly conservative and tends to converge to worse optima. Second, the AP baseline for $perc = 100$ is surprisingly effective, outperforming cEI. Third, using the value of $y(\cdot)$ in the unfeasible region, where the (unfeasible) global optimum resides, degrades performance for cMES. Finally, \figref{results_ranking} shows that cMES ($p=0.9$) is particularly efficient in early iterations, outperforming all competing methods by a wide margin. Individual results for each of the 10 optimization problems are given in Figure~\ref{fig:individual_results} of the supplemental material. We also compared all previous methods as well as PESC in the standard real-valued feedback, observed-objective scenario. The results over the 10 problems are summarized in Figure~\ref{fig:results_ranking_cont} and Table~\ref{tab:results_cont}, showing that cMES outperforms competing approaches.

\begin{table*}[t]
\centering
\begin{tabular}{llcccc}
\toprule
Model                 & Dataset    & Constraint       & Threshold   & Feasible points & $d$ \\
\midrule
XGBoost               & mg         & model size       & 50000 bytes & 72\% & 7 (5, 2, 0)      \\
Decision tree         & mpg        & model size       & 3500 bytes  & 48\% & 4  (2, 1, 1)         \\
Random forest         & pyrim      & model size       & 5000 bytes  & 26\% & 4 (1, 2, 1)        \\
Random forest         & cpusmall   & model size       & 27000 bytes & 80\% & 4(1, 2, 1)           \\
MLP                   & pyrim      & model size       & 27000 bytes & 79\% & 11 (5, 5, 1)         \\
kNN + rnd. projection             & australian & model size       & 28000 bytes & 29\% & 5 (1, 1, 3)          \\
\midrule
MLP                   & heart      & error on neg.\ & 13.3\%      & 30\% & 12  (6, 5, 1)          \\
MLP                   & higgs      & error on neg.\ & 60\%        & 38\% & 12  (6, 5, 1)         \\
Factorization machine & heart      & error on neg.\ & 17\%        & 39\% & 7 (3, 3, 1)         \\
MLP                   & diabetes   & error on neg.\ & 80\%        & 74\% & 12  (6, 5, 1)        \\
\bottomrule
\end{tabular}
\caption{\label{tab:blackbox-description}
   Constrained HPO problems in our experiments. See text for more details. 
   }
\end{table*}

\begin{table}[H]
\centering
\begin{tabular}{lll}
\toprule
       Optimizers & Unfeasible fraction &    Ranking avg \\
\midrule
           $cMES$ &               46.75 &  \textbf{3.08} \\
            $cEI$ &               33.01 &           3.43 \\
 $cMES_{observe}$ &               53.31 &           3.63 \\
  $cEI_{observe}$ &               40.44 &           3.26 \\
             $AP$ &      27.87 &           3.38 \\
         $Random$ &               49.83 &           4.21 \\
\bottomrule
\end{tabular}

\caption{\label{tab:results}
 Binary feedback. Aggregated results for the competing constrained HPO methods.}
\end{table}

\begin{table}[H]
\centering
\begin{tabular}{lll}
\toprule
       Optimizers & Unfeasible fraction &    Ranking avg \\
\midrule
 $cMES_{observe}$ &               44.59 &  \textbf{2.68} \\
  $cEI_{observe}$ &               40.84 &            3.0 \\
 $PESC_{observe}$ &               48.96 &           3.09 \\
             $AP$ &     27.87 &           2.79 \\
         $Random$ &               49.83 &           3.45 \\
\bottomrule
\end{tabular}

\caption{\label{tab:results_cont}
 Real-valued feedback. Aggregated results for the competing constrained HPO methods.}
\end{table}

\begin{figure}[H]
\begin{center}
\centerline{
\includegraphics[width=0.4\textwidth]{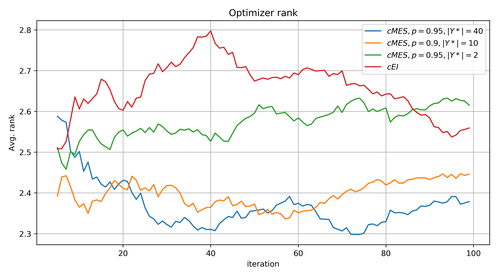}
}
\caption{Binary feedback, unobserved objective. Impact of the number of drawn samples $|Y^*|$.}
\label{figure-ranking-ystar-unobserved}
\end{center}
\end{figure}

\begin{figure}[H]
\begin{center}
\centerline{
\includegraphics[width=0.4\textwidth]{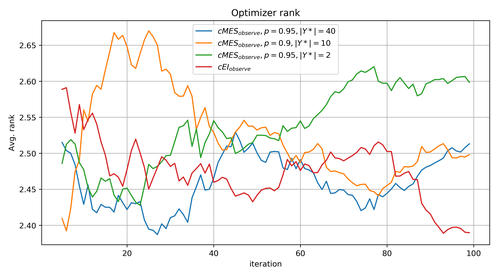}
}
\caption{Binary feedback, observed objective. Impact of the number of drawn samples $|Y^*|$.}
\label{figure-ranking-ystar-observed}
\end{center}
\end{figure}

\begin{table}[H]
\centering
\begin{tabular}{lll}
\toprule
                                 Optimizers &   Marginal    & Joint  \\
                                  & avg rank & avg rank \\
\midrule
$cMES, p=0.1$ &                6.52 & \textbf{6.05} \\
$cMES, p=0.5$ &                     6.1 &  \textbf{5.77} \\
$cMES, p=0.9$ &                       6.21 & \textbf{5.35} \\
 $cMES_{observe}, p=0.1$ &                        7.24 & \textbf{7.04} \\
 $cMES_{observe}, p=0.5$ &                      7.03 &  \textbf{6.85} \\
 $cMES_{observe}, p=0.9$ &                       7.36 &  \textbf{6.45} \\
\bottomrule
\end{tabular}
\caption{\label{tab:marginal}
Binary feedback. Performance of cMES under joint and marginal sampling.}
\end{table}

\vfill

\begin{table}[H]
\centering

\begin{tabular}{lll}
\toprule
              Optimizers &  Avg rank \\
\midrule
  $cMES, p=0.5, |Y*|=40$ &                              5.55 \\
  $cMES, p=0.9, |Y*|=40$ &                           5.5 \\
 $cMES, p=0.95, |Y*|=40$ &                               5.22 \\
  $cMES, p=0.1, |Y*|=10$ &                              5.58 \\
  $cMES, p=0.5, |Y*|=10$ &                               5.48 \\
  $cMES, p=0.9, |Y*|=10$ &                  \textbf{5.09} \\
   $cMES, p=0.5, |Y*|=2$ &                             5.59 \\
   $cMES, p=0.9, |Y*|=2$ &                                   5.6 \\
  $cMES, p=0.95, |Y*|=2$ &                                    5.49 \\
                   $cEI$ &                      5.89 \\
\bottomrule
\end{tabular}
\caption{\label{tab:ystar-unobserved}
Binary feedback, unobserved objective.  Impact of the number of drawn samples $|Y^*|$.}
\end{table}

\vfill

\begin{table}[H]
\centering

\begin{tabular}{lll}
\toprule
                        Optimizers &  Avg rank \\
\midrule
  $cMES_{observe}, p=0.5, |Y*|=40$ &                                5.34 \\
  $cMES_{observe}, p=0.9, |Y*|=40$ &                    5.73 \\
 $cMES_{observe}, p=0.95, |Y*|=40$ &                        5.33 \\
  $cMES_{observe}, p=0.1, |Y*|=10$ &            5.87 \\
  $cMES_{observe}, p=0.5, |Y*|=10$ &                               5.7 \\
  $cMES_{observe}, p=0.9, |Y*|=10$ &                                   5.4 \\
   $cMES_{observe}, p=0.5, |Y*|=2$ &                             \textbf{5.2} \\
   $cMES_{observe}, p=0.9, |Y*|=2$ &                                     5.68 \\
  $cMES_{observe}, p=0.95, |Y*|=2$ &                              5.37 \\
                   $cEI_{observe}$ &                      5.4 \\
\bottomrule
\end{tabular}
\caption{\label{tab:ystar-observed}
Binary feedback, observed objective. Impact of the number of drawn samples $|Y^*|$.}
\end{table}

All experiments with cMES draw the constrained optimum $y_{\star}$ via joint sampling as described in Section~\ref{sec:mes-constraints}. To gain more insight into the ``mean field'' assumption of \cite{Wang:17}, we reran cMES on the 10 constrained optimization problems using their marginal sampling approach to draw $y_{\star}$. The average rankings are reported in Table~\ref{tab:marginal}, where we draw 10 samples of $y_{\star}$  at each iteration either via marginal or joint sampling, both in the observed and unobserved-objective settings and a range of values of $p$. It is clear that marginal sampling degrades optimization performance across the board, confirming our observations from Section~\ref{sec:mes-constraints}. 

We also studied the impact of drawing an increasing number of $y_{\star}$ samples. Let $Y^*$ be a set of all sampled minima, and let $|Y^*|$ be its size. In our experiments, using more than 10 samples of  $y^*$ does not lead to improvement of the algorithm performance. Results are summarized in Figure \ref{figure-ranking-ystar-unobserved} and \ref{figure-ranking-ystar-observed} and Table \ref{tab:ystar-unobserved} and \ref{tab:ystar-observed}.

\section{Conclusion}
\label{sec:conclusion}

In this work, we introduced cMES, a novel acquisition function for Bayesian optimization in the presence of unknown constraints. Our proposed acquisition function can be used both with real-valued and binary constraint feedback. The binary case is relevant in practice, yet underserved in prior work. In an empirical comparison over a wide range of real-world HPO problems, cMES was shown to outperform baselines.

In future work, we will explore modalities where constraints can be evaluated independently of the objective. While cEI alone cannot be used to decide whether to evaluate $y(\xb)$ or $c(\xb)$ in isolation, as shown in our supplemental material it is easy to extend cMES to the separate evaluation case. Experiments with this mode as well as with multiple constraint functions are open directions for future work. We would also like to be able to more directly control the ratio of unfeasible evaluations, which requires changes to the policy beyond the acquisition function. Given our findings about shortcomings of the ``mean field'' independence assumption made in MES \cite{Wang:17}, it could also be important to find alternatives for posterior sampling of the constrained optimum $y_{\star}$. While joint posterior sampling works well in our experiments, its cubic scaling prevents usage for high-dimensional BO search spaces.

\small
\bibliographystyle{unsrt}

\bibliography{bibliography,./template/papers,./template/books,./template/amazon}

\begin{thebibliography}{10}

\bibitem{Mockus1978}
Jonas Mockus, Vytautas Tiesis, and Antanas Zilinskas.
\newblock The application of {B}ayesian methods for seeking the extremum.
\newblock {\em Towards Global Optimization}, 2(117-129):2, 1978.

\bibitem{Jones1998}
Donald~R Jones, Matthias Schonlau, and William~J Welch.
\newblock Efficient global optimization of expensive black-box functions.
\newblock {\em Journal of Global optimization}, 13(4):455--492, 1998.

\bibitem{Shahriari2016}
Bobak Shahriari, Kevin Swersky, Ziyu Wang, Ryan~P Adams, and Nando de~Freitas.
\newblock Taking the human out of the loop: A review of {B}ayesian
  optimization.
\newblock {\em Proceedings of the IEEE}, 104(1):148--175, 2016.

\bibitem{Srinivas:12}
N.~Srinivas, A.~Krause, S.~Kakade, and M.~Seeger.
\newblock Information-theoretic regret bounds for {Gaussian} process
  optimization in the bandit setting.
\newblock {\em IEEE Transactions on Information Theory}, 58:3250--3265, 2012.

\bibitem{Gardner14}
Jacob Gardner, Matt Kusner, Zhixiang Xu, Kilian Weinberger, and John
  Cunningham.
\newblock {Bayesian} optimization with inequality constraints.
\newblock In {\em Proceedings of the International Conference on Machine
  Learning (ICML)}, pages 937--945, 2014.

\bibitem{Gelbart14}
Michael~A. Gelbart, Jasper Snoek, and Ryan~P. Adams.
\newblock {Bayesian} optimization with unknown constraints.
\newblock In {\em Proceedings of the Thirtieth Conference on Uncertainty in
  Artificial Intelligence (UAI)}, pages 250--259, 2014.

\bibitem{Lobato15a}
Jos{\'{e}}~Miguel Hern{\'{a}}ndez{-}Lobato, Michael~A. Gelbart, Matthew~W.
  Hoffman, Ryan~P. Adams, and Zoubin Ghahramani.
\newblock Predictive entropy search for {Bayesian} optimization with unknown
  constraints.
\newblock In {\em Proceedings of the International Conference on Machine
  Learning (ICML)}, pages 1699--1707, 2015.

\bibitem{Hennig:12}
P.~Hennig and C.~Schuler.
\newblock Entropy search for information-efficient global optimization.
\newblock {\em Journal of Machine Learning Research}, 13:1809--1837, 2012.

\bibitem{Hernandez:16}
D.~Hernandez-Lobato, J.~Hernandez-Lobato, A.~Shah, and R.~Adams.
\newblock Predictive entropy search for multi-objective {Bayesian}
  optimization.
\newblock In M.~Balcan and K.~Weinberger, editors, {\em International
  Conference on Machine Learning 33}. JMLR.org, 2016.

\bibitem{Lobato2016}
Jos\'{e}~Miguel Hern\'{a}ndez-Lobato, Michael~A. Gelbart, Ryan~P. Adams,
  Matthew~W. Hoffman, and Zoubin Ghahramani.
\newblock A general framework for constrained bayesian optimization using
  information-based search.
\newblock {\em Journal of Machine Learning Research}, 17(160):1--53, 2016.

\bibitem{Wang:17}
Z.~Wang and S.~Jegelka.
\newblock Max-value entropy search for efficient {Bayesian} optimization.
\newblock In D.~Precup and Y.~W. Teh, editors, {\em International Conference on
  Machine Learning 34}. JMLR.org, 2017.

\bibitem{Eggensperger2012}
K~Eggensperger, F~Hutter, HH~Hoos, and K~Leyton-brown.
\newblock Efficient benchmarking of hyperparameter optimizers via surrogates
  background: Hyperparameter optimization.
\newblock In {\em Proceedings of the 29th AAAI Conference on Artificial
  Intelligence (AAAI)}, pages 1114--1120, 2012.

\bibitem{Rasmussen2006}
Carl Rasmussen and Chris Williams.
\newblock {\em {Gaussian} Processes for Machine Learning}.
\newblock MIT Press, 2006.

\bibitem{Snoek2015}
Jasper Snoek, Oren Rippel, Kevin Swersky, Ryan Kiros, Nadathur Satish,
  Narayanan Sundaram, Mostofa Patwary, Mr~Prabhat, and Ryan Adams.
\newblock Scalable {B}ayesian optimization using deep neural networks.
\newblock In {\em Proceedings of the International Conference on Machine
  Learning (ICML)}, pages 2171--2180, 2015.

\bibitem{letham2019}
Benjamin Letham, Brian Karrer, Guilherme Ottoni, and Eytan Bakshy.
\newblock Constrained bayesian optimization with noisy experiments.
\newblock {\em Bayesian Analysis}, 14(2):495--519, 2019.

\bibitem{Picheny2016}
V.~Picheny, R.~B. Gramacy, S.~Wild, and S.~Le~Digabel.
\newblock {Bayesian optimization under mixed constraints with a slack-variable
  augmented Lagrangian}.
\newblock {\em Advances in Neural Information Processing Systems 29}, 2016.

\bibitem{Ariafar2019}
Setareh Ariafar, Jaume Coll-Font, Dana Brooks, and Jennifer Dy.
\newblock Admmbo: Bayesian optimization with unknown constraints using admm.
\newblock {\em Journal of Machine Learning Research}, 20(123):1--26, 2019.

\bibitem{Gpyopt2016}
{GPyOpt}: A {B}ayesian optimization framework in {Python}.
\newblock \texttt{http://github.com/SheffieldML/GPyOpt}, 2016.

\bibitem{Minka:01a}
T.~Minka.
\newblock Expectation propagation for approximate {Bayesian} inference.
\newblock In J.~Breese and D.~Koller, editors, {\em Uncertainty in Artificial
  Intelligence 17}. Morgan Kaufmann, 2001.

\bibitem{Bishop:06}
C.~Bishop.
\newblock {\em Pattern Recognition and Machine Learning}.
\newblock Springer, 1st edition, 2006.

\bibitem{Thompson:33}
W.~Thompson.
\newblock On the likelihood that one unknown probability exceeds another in
  view of theevidence of two samples.
\newblock {\em Biometrika}, 25(3):285--294, 1933.

\bibitem{Kandasamy:18}
K.~Kandasamy, A.~Krishnamurthy, J.~Schneider, and B.~Poczos.
\newblock Parallelised {Bayesian} optimisation via {Thompson} sampling.
\newblock In A.~Gretton and C.~Robert, editors, {\em Workshop on Artificial
  Intelligence and Statistics 19}, pages 133--142, 2016.

\bibitem{Sobol1967}
Ilya~M Sobol.
\newblock On the distribution of points in a cube and the approximate
  evaluation of integrals.
\newblock {\em USSR Computational Mathematics and Mathematical Physics},
  7(4):86--112, 1967.

\bibitem{Bergstra2012}
James Bergstra and Yoshua Bengio.
\newblock Random search for hyper-parameter optimization.
\newblock {\em Journal of Machine Learning Research (JMLR)}, 13:281--305, 2012.

\bibitem{pedregosa2011scikit}
Fabian Pedregosa, Ga\"{e}l Varoquaux, Alexandre Gramfort, Vincent Michel,
  Bertrand Thirion, Olivier Grisel, Mathieu Blondel, Peter Prettenhofer, Ron
  Weiss, Vincent Dubourg, Jake Vanderplas, Alexandre Passos, David Cournapeau,
  Matthieu Brucher, Matthieu Perrot, and \'{E}douard Duchesnay.
\newblock Scikit-learn: Machine learning in {Python}.
\newblock {\em Journal of Machine Learning Research (JMLR)}, 12:2825--2830,
  2011.

\bibitem{Chang2011}
Chih-Chung Chang and Chih-Jen Lin.
\newblock {LIBSVM}: A library for support vector machines.
\newblock {\em ACM Transactions on Intelligent Systems and Technology},
  2:27:1--27:27, 2011.

\bibitem{Feurer:15}
M.~Feurer, A.~Klein, K.~Eggensperger, J.~Springenberg, M.~Blum, and F.~Hutter.
\newblock Efficient and robust automated machine learning.
\newblock In {\em Advances in Neural Information Processing Systems 28}, pages
  2962--2970, 2015.

\bibitem{Bardenet2013}
R.~Bardenet, M.~Brendel, B.~K{\'e}gl, and M.~Sebag.
\newblock Collaborative hyperparameter tuning.
\newblock In {\em Proceedings of the International Conference on Machine
  Learning (ICML)}, pages 199--207, 2013.

\bibitem{klein19}
A.~Klein, Z.~Dai, F.~Hutter, N.~Lawrence, and J.~Gonzalez.
\newblock Meta-surrogate benchmarking for hyperparameter optimization.
\newblock {\em arXiv:1905.12982}, 2019.

\end{thebibliography}

\clearpage
\appendix

\onecolumn
\section*{Supplementary material}

\section{Derivations}
\label{sec:derivations}

Consider the problem of Bayesian Optimization (BO) with unknown constraints. There are two real-valued functions $y(\vx{})$, $c(\vx{})$ over a common space $\mathcal{X}$. The first models the criterion to be minimized, the second parameterizes the constraint. An evaluation produces $z_y$, $z_c$ according to likelihood functions. First,
\[
  z_y\sim N(z_y | y(\vx{}), \alpha_y^{-1}).
\]
For $z_c$, we consider two different options. In one scenario, we may observe $c(\vx{})$ directly, up to Gaussian noise:
\[
  z_c\sim N(z_c | c(\vx{}), \alpha_c^{-1}).
\]
In a different scenario, we may observe a binary target only:
\[
  z_c\sim \sigma(z_c c(\vx{})),\quad z_c\in\{\pm 1\}.
\]
Here, $z_c=-1$ means the evaluation at $\vx{}$ is feasible, and $z_c=+1$ means it is infeasible. We use the logistic parameterization, involving
\[
  \sigma(t) = \frac{1}{1 + e^{-t}},
\]
but any other likelihood could be used instead. The constrained optimization problem we would like to solve is
\begin{equation}\label{eq:ystar2}
  y_* = \min_{\vx{}\in\mathcal{X}}\left\{ y(\vx{})\; \|\; c(\vx{}) \le \delta \right\}.
\end{equation}
Here, $\delta$ is a confidence parameter. In the case of binary feedback, $z_c\in\{\pm 1\}$, we can also write
\[
  y_* = \min_{\vx{}\in\mathcal{X}}\left\{ y(\vx{})\; \|\; P(z_c = +1 | \vx{}) =
  \sigma(c(\vx{})) \le \sigma(\delta) \right\},
\]
where the confidence parameter is $\sigma(\delta)\in (0,1)$. Importantly, both $y(\cdot)$ and $c(\cdot)$ are unknown up front and have to be learned from noisy samples $z_y, z_c$.

We assume that some data $\mathcal{D}$ has already been acquired, based on which independent {\em Gaussian} posterior processes are obtained for $y(\vx{})$ and $c(\vx{})$. The marginals of these are denoted by $N(y | \mu_y, \sigma_y^2)$ and $N(c | \mu_c, \sigma_c^2)$, where we drop the indexing by $\vx{}$. In the sequel, we drop both the conditioning on $\vx{}$ and on $\mathcal{D}$ from the notation. For example, we write $P(y, c)$ instead of $P(y, c | \mathcal{D}, \vx{})$:
\[
  P(y, c) = P(y) P(c) = N(y | \mu_y, \sigma_y^2) N(c | \mu_c, \sigma_c^2).
\]

The MES acquisition function \cite{Wang:17} without constraints is given by:
\[
  \mathcal{I}( y ; y_* ) = \Ent[ P(y) ] - \Ex\left[ \Ent[ P(y | y_*) ] \right],
\]
where the expectation is over $P(y_* | \mathcal{D})$, and $y_* = \min_{\vx{}\in\mathcal{X}} y(\vx{})$. Here, $P(y | y_*) \propto P(y) \Ind{y\ge y_*}$ is a truncated Gaussian. It should be noted that this is a simplifying assumption. In PES \cite{Hernandez:16}, the related distribution $P(y | \vx{*})$ is approximated, where $\vx{*}$ is the argmin. Several local constraints on $y(\cdot)$ at $\vx{*}$ are taken into account, such as $\nabla_{\vx{*}} y = \vzero$. This is not done in MES, which simplifies derivations dramatically. Second, the expectation over $y_*$ is approximated by Monte Carlo sampling.

\subsection{Real-valued Constraint Feedback}
\label{sec:real-valued-feedback}

In this section, we assume that the constraint function $c(\cdot)$ can be observed directly, so that we obtain real-valued feedback from both $y(\cdot)$ and $c(\cdot)$. Our generalization of MES to the constrained case uses
\begin{equation}\label{eq:afunc-nonoise}
  A_1(\vx{}) = \mathcal{I}( (y, c) ; y_* ) = \Ent[ P(y, c) ] - \Ex\left[ \Ent[ P(y, c | y_*) ]
  \right],
\end{equation}
where the expectation is over $P(y_* | \mathcal{D})$, and $y_*$ is the constrained minimum \eqp{ystar}. There are two points to be worked out:
\begin{itemize}
\item
  Expression $\Ent[ P(y, c) ] - \Ent[ P(y, c | y_*) ]$ for fixed $y_*$
\item
  Efficient approximate sampler from $P(y_* | \mathcal{D})$, where $y_*$ is given
  by \eqp{ystar}, given that $y(\cdot)$ and $c(\cdot)$ are sampled from their
  respective posterior distributions (assumed to be independent).
\end{itemize}

In fact, the formulation so far ignores that we observe $z_y, z_c$ at $\vx{}$, not $y(\vx{})$, $c(\vx{})$. Even though this is ignored in the original MES paper, a better acquisition function would therefore be
\begin{equation}\label{eq:afunc-noise}
  A_2(\vx{}) = \mathcal{I}( (z_y, z_c) ; y_* ) = \Ent[ P(z_y, z_c) ] -
  \Ex\left[ \Ent[ P(z_y, z_c | y_*) ] \right].
\end{equation}

We start with the entropy difference in \eqp{afunc-nonoise}, where noise models are ignored, and come back to the noisy case \eqp{afunc-noise} below. We will define $P(y, c | y_*)$ in the same ``local'' way as in MES, avoiding all complications as considered in PES. What do we learn by conditioning on $y_*$? If $c \le \delta$, then $y\ge y_*$. Otherwise ($c>\delta$), our belief in $y$ remains the same. Therefore:
\[
  P(y, c | y_*) = Z^{-1} P(y, c) \Ind{c > \delta \lor y\ge y_*} = Z^{-1} P(y, c)
  (1 - \Ind{c\le\delta} \Ind{y\le y_*}).
\]
Here, we replaced $y<y_*$ by $y\le y_*$, which makes no difference for a distribution with a density. In the remainder of this section, $\Ex[\cdot]$ is always over $P(y, c)$, unless otherwise indicated. Denote
\[
  \kappa(y, c) := 1 - \Ind{c\le\delta} \Ind{y\le y_*}\quad\Rightarrow\quad
  P(y, c | y_*) = Z^{-1} P(y, c) \kappa(y, c).
\]

We need some notation:
\[
  \gamma_c := \frac{\delta - \mu_c}{\sigma_c},\quad \gamma_y := \frac{y_* -
  \mu_y}{\sigma_y},\quad Z_c = \Ex[\Ind{c\le\delta}] = \Phi(\gamma_c),\quad
  Z_y = \Ex[\Ind{y\le y_*}] = \Phi(\gamma_y).
\]
Here, $\Phi(t) = \Ex[\Ind{n\le t}]$, $n\sim N(0,1)$, is the cumulative distribution function for a standard normal variate. The normalization constant is
\[
  Z = \Ex[\kappa(y, c)] = 1 - Z_c Z_y.
\]
Also,
\[
  \Ent[ P(y, c | y_*) ] = Z^{-1}\Ex\left[ \kappa(y, c) ( \log Z - \log P(y, c) ) \right] =
  \log Z + Z^{-1} \Ex\left[ \kappa(y, c) (-\log P(y, c) ) \right].
\]
Note that the $-\log\kappa(y, c)$ drops out, because $1\log 1 = 0\log 0 = 0$. If we parameterize $c = \mu_c + \sigma_c n_c$, $y = \mu_y + \sigma_y n_y$, where $n_c, n_y$ are independent $N(0, 1)$ variates, we have that
\[
  -\log P(y, c) = \frac{1}2\left( n_c^2 + n_y^2 + \log(2\pi\sigma_c^2) +
  \log(2\pi\sigma_y^2) \right).
\]
Plugging this in:
\[
  \Ent[ P(y, c | y_*) ] = \log Z + \frac{1}2\left( \log(2\pi\sigma_c^2) +
  \log(2\pi\sigma_y^2) \right) + \frac{1}{2 Z} \Ex\left[ (1 - \Ind{n_c\le\gamma_c}
  \Ind{n_y\le\gamma_y}) (n_c^2 + n_y^2) \right].
\]
At this point, we need the simple identity:
\[
  \Ex[\Ind{n\le\gamma} n^2] = \Ex[\Ind{n\le\gamma}] - \gamma N(\gamma) =
  \Phi(\gamma) - \gamma N(\gamma),\quad N(x) := N(x | 0, 1).
\]
Concentrating on the final expectation term:
\[
\begin{split}
  & (2 Z)^{-1}\Ex[\dots] = Z^{-1} - (2 Z)^{-1} \Ex\left[ \Ind{n_c\le\gamma_c}
  \Ind{n_y\le\gamma_y} (n_c^2 + n_y^2) \right] \\
  & = Z^{-1} - (2 Z)^{-1}\left( Z_y
  (Z_c - \gamma_c N(\gamma_c)) + Z_c (Z_y - \gamma_y N(\gamma_y)) \right)
  = Z^{-1}\left( Z + \frac{1}2\left( Z_y \gamma_c N(\gamma_c) + Z_c \gamma_y
  N(\gamma_y) \right) \right).
\end{split}
\]
The hazard function of the standard normal is defined as
\[
  h(x) := \frac{N(x)}{\Phi(-x)}.
\]
Noting that $\Ent[ P(y, c) ] = \Ent[ P(y) ] + \Ent[ P(c) ]$ and $\Ent[
P(y) ] = (1 + \log(2\pi\sigma_y^2))/2$, some algebra gives
\[
  \Ent[ P(y, c | y_*) ] = \Ent[ P(y, c) ] + \log Z + \frac{\gamma_c h(-\gamma_c) +
  \gamma_y h(-\gamma_y)}{2 (\exp(-\log Z_c -\log Z_y) - 1)}.
\]
Here, we used
\[
  \frac{Z_y Z_c}{Z} = \frac{Z_y Z_c}{1 - Z_y Z_c} =
  \frac{1}{\exp(-\log Z_c -\log Z_y) - 1}.
\]
All in all:
\[
  \Ent[ P(y, c) ] - \Ent[ P(y, c | y_*) ] = -\log Z - \frac{\gamma_c h(-\gamma_c) +
  \gamma_y h(-\gamma_y)}{2 (\exp(-\log Z_c -\log Z_y) - 1)}.
\]
Note that $\log Z_c$, $\log Z_y$ are negative. The only case when this expression becomes problematic is if both $\log Z_y$ and $\log Z_c$ tend to zero. This happens only if both $y$ is much smaller than $y_*$ and $c$ is much smaller than $\delta$. If $y_*$ is sampled from $P(y_* | \mathcal{D})$, this is very unlikely to be the case.
We need numerically robust code for computing $\log\Phi(x)$ and $h(x)$.

\subsection{Entropy Difference for Noisy Targets}
\label{sec:gaussian-noise}

As noted above, we would ideally compute the entropy difference for the noisy targets $z_y, z_c$ instead of the latents $y, c$, so use the acquisition function \eqp{afunc-noise} instead of \eqp{afunc-nonoise}. How would this look like for the case where both $z_y$ and $z_c$ are real-valued with Gaussian likelihood? Define
\[
  \Psi(z_y, z_c) = \int P(z_y, y) P(z_c, c) \kappa(y, c)\, d y d c = P(z_y) P(z_c) \left(
  1 - \tilde{Z}_y(z_y) \tilde{Z}_c(z_c) \right),
\]
where $\tilde{Z}_y(z_y)$ is defined as $Z_y$, but with $P(y)$ being replaced by the posterior $P(y | z_y)$. Then:
\[
  P(z_y, z_c | y_*) = Z^{-1} \Psi(z_y, z_c),\quad Z = 1 - Z_y Z_c.
\]
To our knowledge, there is no simple closed-form expression for $\Ent[P(z_y, z_c | y_*)]$. The problem is that $\Psi(z_y, z_c)$ is not the product of a Gaussian with an indicator, and in particular $\log \Psi(z_y, z_c)$ is a complex function.

Here is a simple idea which may work better than just ignoring the noise and using \eqp{afunc-nonoise}. Complications arise because the expectations over $P(y | z_y)$ and $P(c | z_c)$ in $\Psi(z_y, z_c)$ do not result in a term which is the product of Gaussians and indicators. We can mitigate this problem by approximating $P(y | z_y)$ with $\delta(y - \Ex[y | z_y])$. Doing so results in
\[
  \Psi(z_y, z_c) = P(z_y) P(z_c) (1 - \Ind{\Ex[y | z_y]\le y_*} \Ind{\Ex[c | z_c]\le \delta}).
\]
Here, $P(z_y) = N(\mu_y, \sigma_y^2 + \alpha_y^{-1})$, $P(z_c) = N(\mu_c, \sigma_c^2 + \alpha_c^{-1})$. Since $\Ex[y | z_y]$ is an affine function of $z_y$, this can be brought into the same form as is used in the noise-free case, but $y$ is replaced by $z_y$, $y_*$ by a different value, and $P(y)$ by $P(z_y)$. Namely,
\[
  \Ex[y | z_y] = \mu_y + \frac{\sigma_y^2}{\sigma_y^2 + \alpha_y^{-1}} (z_y - \mu_y)
  = \mu_y + \rho_y^2 (z_y - \mu_y),\quad \rho_y^2 =
  \frac{\sigma_y^2\alpha_y}{1 + \sigma_y^2\alpha_y},
\]
so that
\[
  \Ex[y | z_y]\le y_*\quad \Leftrightarrow\quad z_y \le \tilde{y}_* := \mu_y +
  \rho_y^{-2}(y_* - \mu_y).
\]
We can now simply use the derivation from above. In fact,
\[
  \tilde{\gamma}_y = \frac{\tilde{y}_* - \mu_y}{(\sigma_y^2 + \alpha_y^{-1})^{1/2}}
  = \frac{y_* - \mu_y}{\sigma_y \rho_y},\quad \tilde{\gamma}_c =
  \frac{\delta - \mu_c}{\sigma_c \rho_c},\quad \rho_y =
  \frac{\sigma_y\alpha_y^{1/2}}{\sqrt{ 1 + (\sigma_y\alpha_y^{1/2})^2}}
\]
just have to be used instead of $\gamma_y, \gamma_c$.

\subsection{Binary Constraint Feedback}
\label{sec:entdiff-binary}

For binary response $z_c\in\{\pm 1\}$, we have to take into account that much less information is obtained by sampling the constraint at $\vx{}$. Here, a sensible approach is to ignore the noise on $y$, but not ignore the likelihood $c\to z_c$. In other words, we can try to approximate
\begin{equation}\label{eq:afunc-mixed}
  A_3(\vx{}) = \mathcal{I}( (y, z_c) ; y_* ) = \Ent[ P(y, z_c) ] -
  \Ex\left[ \Ent[ P(y, z_c | y_*) ] \right].
\end{equation}
In this case, we use some approximate inference method for
\[
  Q(z_c) Q(c | z_c) \approx P(z_c | c) P(c),\quad z_c\in\{\pm 1\},
\]
where $Q(c | z_c)$ are Gaussians. In our current code, we use Laplace's approximation, where mode finding is approximated by a single Newton step. Also, $Q(z_c)$ is using the highly accurate approximation given in \cite[Sect.~4.5.2]{Bishop:06}. Now:
\[
\begin{split}
  & \Psi(y, z_c) := \int Q(z_c) Q(c | z_c) P(y) \kappa(y, c)\, d c = P(y) Q(z_c)
  \tskappa{}(y, z_c), \\
  & \tskappa{}(y, z_c) := \left( 1 - \Ind{y\le y_*} F(z_c) \right),\; F(z_c) =
  \Ex_{Q(c|z_c)}[\Ind{c\le\delta}],
\end{split}
\]
and
\[
  P(y, z_c | y_*) \approx Z^{-1}\Psi(y, z_c),\quad Z = 1 - Z_y \tilde{Z}_c,\quad
  \tilde{Z}_c = \Ex_Q[F(z_c)].
\]
Importantly, $\tskappa{}(y, z_c)$ is piece-wise constant, while not an indicator function anymore. Note that $\tilde{Z}_c\ne Z_c$ in general, due to the approximation we use, but it should be close.

In the following, $\Ex[\cdot]$ is over $P(y) Q(z_c)$, $\Ex_P[\cdot]$ is over $P(y)$, and $\Ex_Q[\cdot]$ is over $Q(z_c)$. First,
\[
\begin{split}
  & \Ent[P(y, z_c | y_*)] = \log Z + Z^{-1}\Ex\left[ \tskappa{}(y, z_c) \left( -\log P(y)
  -\log Q(z_c) - \log \tskappa{}(y, z_c) \right) \right] \\
  & = \log Z + \frac{1}2\log(2\pi\sigma_y^2) + \Ex_Q[G(z_c)], \\
  & G(z_c) := Z^{-1}\Ex_P\left[ \tskappa{}(y, z_c) \left( n_y^2/2 -\log Q(z_c)
  - \log \tskappa{}(y, z_c) \right) \right].
\end{split}
\]
We split this in three parts, using the derivation of the noise-free case above. First:
\[
  G_1(z_c) = Z^{-1}\Ex_P\left[ \tskappa{}(y, z_c) n_y^2/2 \right] = \frac{1}{2 Z}\left(
  1 - F(z_c) (Z_y - \gamma_y N(\gamma_y)) \right).
\]
Next:
\[
  G_2(z_c) = Z^{-1}\Ex_P\left[ \tskappa{}(y, z_c) (-\log Q(z_c)) \right] = Z^{-1}
  (1 - Z_y F(z_c)) (-\log Q(z_c)).
\]
Finally, note that if $y\ge y_*$, then $\log\tskappa{}(y, z_c) = \log 1 = 0$, so we can replace $\tskappa{}(y, z_c)$ by $\Ind{y\le y_*} (1 - F(z_c))$, therefore:
\[
  G_3(z_c) = Z^{-1}\Ex_P\left[ \tskappa{}(y, z_c) ( -\log \tskappa{}(y, z_c) ) \right]
  = Z^{-1} Z_y (1 - F(z_c)) (-\log(1 - F(z_c))).
\]
Next, the expectation over $Q(z_c)$. First,
\[
  G_1(z_c) = \frac{1}{2 Z}\left( 1 - F(z_c) Z_y + F(z_c) \gamma_y N(\gamma_y)
  \right),
\]
so that
\[
  \Ex_Q[G_1(z_c)] = \frac{1}2 + \frac{Z_y \tilde{Z}_c}{2 Z} \gamma_y h(-\gamma_y).
\]
Next, using $1 - Z_y F(z_c) = Z - Z_y(F(z_c) - \tilde{Z}_c)$:
\[
\begin{split}
  \Ex_Q[G_2(z_c)] & = Z^{-1}\Ex_Q\left[ (1 - Z_y F(z_c)) (-\log Q(z_c)) \right] \\
  & = \Ent[Q(z_c)] - \frac{Z_y\tilde{Z}_c}{Z} \tilde{Z}_c^{-1} \Ex_Q\left[ (F(z_c) -
  \tilde{Z}_c) (-\log Q(z_c)) \right].
\end{split}
\]
Finally,
\[
  \Ex_Q[G_3(z_c)] = \frac{Z_y\tilde{Z}_c}{Z} \tilde{Z}_c^{-1}\Ex_Q\left[ (1 - F(z_c))
  (-\log(1 - F(z_c))) \right].
\]
Altogether, we obtain
\[
\begin{split}
  & \Ent[P(y)] + \Ent[Q(z_c)] - \Ent[P(y, z_c | y_*)] = -\log Z \\
  & - B\left( \gamma_y h(-\gamma_y)/2 + \tilde{Z}_c^{-1}\Ex_Q\left[
  (1 - F(z_c)) (-\log(1 - F(z_c))) + (F(z_c) - \tilde{Z}_c) \log Q(z_c) \right] \right), \\
  & B = \frac{Z_y \tilde{Z}_c}{Z} = \frac{1}{\exp(-\log Z_y -\log \tilde{Z}_c) - 1}.
\end{split}
\]
We would compute $\log Z_y$, $h(-\gamma_y)$, $\log F(z_c)$, $\log (1-F(z_c))$, then $\log\tilde{Z}_c$ by {\tt logsumexp}. In fact, if
\[
  \gamma_c(z_c) = \frac{\delta - \Ex_Q[c | z_c]}{\sqrt{\Var_Q[c | z_c]}},
\]
then
\[
  \log F(z_c) = \log\Phi(\gamma_c(z_c)),\quad \log (1 - F(z_c)) =
  \log\Phi(-\gamma_c(z_c)).
\]
The term $\tilde{Z}_c^{-1}\Ex_Q[\dots]$ is computed by folding the normalization into the argument inside $\Ex_Q[\dots]$, which is computed as
\[
  \left( e^{\log F(z_c) - \log \tilde{Z}_c} - 1 \right) \log Q(z_c) -
  e^{\log (1 - F(z_c)) - \log \tilde{Z}_c} \log (1 - F(z_c)).
\]
We then multiply with $Q(z_c)$ and sum over $z_c = -1, +1$.

\subsection{Sampling from $P(y_{\star} | \mathcal{D})$}
\label{sec:sampling-ystar}

In the constrained case, we aim to sample from $P(y_{\star} | \mathcal{D})$, where $y_{\star} = \min_{\vx{}\in\mathcal{X}}\left\{ y(\vx{})\; \|\; c(\vx{}) \le \delta \right\}$. Here, $y(\cdot)$ and $c(\cdot)$ are posterior GPs conditioned on the current data $\mathcal{D}$. At least for commonly used infinite-dimensional kernels, it is intractable to draw exact sample functions from these GPs, let alone to solve the conditional optimization problem for $y_{\star}$.

In \cite{Lobato15a}, a finite-dimensional random kitchen sink (RKS) approximation is used to draw approximate sample paths, and the constrained problem is solved for these. Since the RKS basis functions are nonlinear in $\vx{}$, so are objective and constraint function, and solving for $y_{\star}$ requires complex machinery. A simpler approach is used in \cite{Wang:17}. They target the cumulative distribution function (CDF) of $y_{\star}$, which can be written as expectation over $y(\cdot)$ and $c(\cdot)$ of an infinite product. This is approximated by restricting the product over a finite set $\hat{\mathcal{X}}$, and by assuming {\em independence} of all $y(\vx{})$ and $c(\vx{})$ for $\vx{}\in \hat{\mathcal{X}}$. While this gives rise to a tractable approximation of the CDF, we found this approximation to be problematic in our experiments. As noted in \cite{Wang:17}, $y_{\star}$ drawn under these assumptions are underbiased. In fact, due to the independence assumption, this bias gets {\em worse} the larger $\hat{\mathcal{X}}$ is: $y_{\star}$ diverges as $|\hat{\mathcal{X}}|\to \infty$.

In our experiments, we follow \cite{Wang:17} by restricting our attention to a finite set $\hat{\mathcal{X}}$ (we use a Sobol sequence \cite{Sobol1967}), but then draw {\em joint} samples of $y(\hat{\mathcal{X}})$ and $c(\hat{\mathcal{X}})$ respectively, based on which $y_{\star}$ (restricted to $\hat{\mathcal{X}}$) is trivial to compute. While joint sampling scales cubically in the size of $\hat{\mathcal{X}}$, sampling takes less than a second for $|\hat{\mathcal{X}}| = 2000$, the size we used in our experiments.

More precisely, the posterior for $y(\cdot)$ conditioned on data $\vz{y} = [z_{y i}]\in\R^n$ is defined in terms of the Cholesky factor $\mxl{}$ and the vector $\vp{}$, where
\[
   \mxl{}\mxl{}^T = \mxk{} + \alpha_y^{-1}\Id,\quad \vp{} = \mxl{}^{-1}\vz{y},
\]
where $\mxk{} = k_y(\mxx{}, \mxx{})\in\R^{n\times n}$ is the kernel matrix on the training set ($\mxx{} = [\vx{i}]\in\R^{n\times p}]$), and $\alpha_y$ is the noise precision. The posterior distribution of $y(\hat{\mathcal{X}})$ is a Gaussian with mean and covariance
\[
   \hvmu{} = \mxm{}\vp{},\quad \mxm{} = \mxk{*,\cdot}\mxl{}^{-T}, \quad
   \hmxsigma{} = \mxk{*, *} - \mxm{}\mxm{}^T,
\]
where $\mxk{*,\cdot} = k_y(\hat{\mathcal{X}}, \mxx{})\in\R^{m\times n}$, $m = |\hat{\mathcal{X}}|$, and $\mxk{*, *} = k_y(\hat{\mathcal{X}}, \hat{\mathcal{X}})\in \R^{m\times m}$. Samples of $y(\hat{\mathcal{X}})$ are drawn as
\[
  \hmxy{} = \hmxl{}\mxn{} + \hvmu{} \vone_k^T,\quad \hmxl{}\hmxl{}{}^T =
  \hmxsigma{}, \quad \mxn{} = [\nu_{r s}]\in\R^{m\times k},\;
  \nu_{r s}\sim N(0, 1).
\]
Due to the Cholesky factorization, joint sampling scales cubically in $m$. On the other hand, sampling takes less than one second for sizes smaller than 2000.

\subsection{Scoring Constraint or Criterion Evaluation}
\label{sec:entdiff-marginal}

In some situations, we may be able to evaluate criterion and constraints independent of each other. For example, one may be much cheaper to evaluate than the other. To this end, we would like to score the value of sampling $y(\vx{})$ or $c(\vx{})$ at $\vx{}$.

To this end, we just marginalize the joint distributions worked out above. First, consider the case where $z_y, z_c$ are real-valued, and we would like to score the value of sampling $z_y$ (the case of sampling $z_c$ is symmetric then). Note that this is the noisy case, where $\sigma_y$ is replaced by $\sigma_y\rho_y$, $\gamma_y$ by $\tsgamma{y}$, etc. We have that
\[
   \Psi(z_y) = P(z_y) ( 1 - \Ind{\Ex[y|z_y]\le y_*} Z_c ),\quad Z = 1 - \tilde{Z}_y Z_c.
\]
Then, $P(z_y | y_*) = Z^{-1} \Psi(z_y)$. Note that $Z_c = \Phi(\gamma_c)$ without the noise. In fact, the marginal does not depend on the noise $c\to z_c$, so the same expression is obtained in the case $z_c\in\{\pm 1\}$. In the following, we use that
\[
  \log ( 1 - \Ind{\Ex[y|z_y]\le y_*} Z_c) = \Ind{\Ex[y|z_y]\le y_*} \log(1 - Z_c).
\]
Then:
\[
\begin{split}
  \Ent[P(z_y | y_*)] & = \log Z + \frac{1}2\log(2\pi\Var[z_y]) + Z^{-1}\Ex\Bigl[
  (1 - Z_c\Ind{n_y\le \tsgamma{y}}) n_y^2/2 \\
  & + \Ind{n_y\le \tsgamma{y}} (1-Z_c) (-\log(1-Z_c)) \Bigr].
\end{split}
\]
Some algebra gives
\[
  \Ent[P(z_y)] - \Ent[P(z_y | y_*)]  = -\log Z - \frac{\tsgamma{y} h(-\tsgamma{y})/2
  - Z_c^{-1}(1-Z_c) \log(1 - Z_c)}{\exp(-\log\tilde{Z}_y -\log Z_c) - 1},\quad Z =
  1 - \tilde{Z}_y Z_c.
\]
By symmetry, if $z_c\in\R$ with Gaussian noise:
\[
  \Ent[P(z_c)] - \Ent[P(z_c | y_*)]  = -\log Z - \frac{\tsgamma{c} h(-\tsgamma{c})/2
  - Z_y^{-1}(1-Z_y) \log(1 - Z_y)}{\exp(-\log\tilde{Z}_c -\log Z_y) - 1},\quad Z =
  1 - Z_y \tilde{Z}_c.
\]

Finally, consider $z_c\in\{\pm 1\}$. Here,
\[
  \tskappa{}(z_c) = 1 - Z_y F(z_c),\quad Z = 1 - Z_y\tilde{Z}_c,\quad P(z_c | y_*) =
  Z^{-1} Q(z_c) \tskappa{}(z_c).
\]
Then:
\[
  \Ent[P(z_c | y_*)] = \log Z + Z^{-1}\Ex_Q\left[ \tskappa{}(z_c) \left( -\log Q(z_c)
  - \log\tskappa{}(z_c) \right) \right].
\]
Some algebra gives
\[
  \Ent[Q(z_c)] - \Ent[P(z_c | y_*)] = -\log Z + Z^{-1}\Ex_Q\left[ \tskappa{}(z_c)
  \log\tskappa{}(z_c) - Z_y(F(z_c) - \tilde{Z}_c) \log Q(z_c) \right],
\]
where $Z = 1 - Z_y \tilde{Z}_c$. Using the notation from above, this can also be written as
\[
\begin{split}
  & \Ent[Q(z_c)] - \Ent[P(z_c | y_*)] = -\log Z - Z^{-1}\Ex_Q[ \tskappa{}(z_c)
  (-\log\tskappa{}(z_c)) ] \\
  & - B \tilde{Z}_c^{-1}\Ex_Q[ (F(z_c) - \tilde{Z}_c) \log Q(z_c) ], \quad
  B = \frac{Z_y \tilde{Z}_c}{Z} = \frac{1}{\exp(-\log Z_y -\log \tilde{Z}_c) - 1}.
\end{split}
\]

\subsection{Observe $y(\vx{})$ only in Feasible Region}

In this section, we deal with binary feedback $z_c\in\{-1,+1\}$. For some important applications, feedback $z_y$ on $y(\vx{})$ is obtained only if $z_c = -1$ (feasible). For example, BO may be used to tune parameters of deep neural networks. A function evaluation $z_y$ of a test set metric may fail, because training crashed due to out of memory errors ($z_c = +1$). Note that $y_*$ itself does not depend on values of $y(\vx{})$ in the infeasible region.

It seems hard to properly define the entropy difference (conditioned on $y_*$) in this case. One idea is to simply use the entropy difference from \secref{entdiff-binary}. Even though this assumes noise-free feedback for $y$, the value conveys information about $y_*$ only if $\vx{}$ is feasible. Another idea is to consider the mixture of $Q(z_c = -1)$ times the entropy difference from \secref{entdiff-binary} plus $Q(z_c = +1)$ times the entropy difference from \secref{entdiff-marginal}. At least for $Q(z_c)$ away from $1/2$, this could be a more reasonable score. Note that the part
\[
   -\log Z - B \tilde{Z}_c^{-1}\Ex_Q[ (F(z_c) - \tilde{Z}_c) \log Q(z_c) ]
\]
appears in both entropy difference expressions.

\section{Real-world Hyperparameter Tuning Problems}
\label{sec:experiment-details}
We considered a range of 10 constrained HPO problems, spanning different {\tt scikitlearn} algorithms \cite{pedregosa2011scikit}, {\tt libsvm} datasets \cite{Chang2011}, and constraint modalities. The first six problems are about optimizing an accuracy metric (AUC for binary classification, coefficient of determination for regression) subject to a constraint on model size, a setup motivated by applications in IOT or on mobile devices. The remaining four problems require minimizing the error on positives, subject to a limit on the error on negatives, as is relevant for example in applications in medical domains; here, one hyperparameter to tune is the fraction of the positive class in the data (both training and validation), which is adjusted by resampling with replacement. A summary of algorithms, datasets, and fraction of feasible configurations is given in \tabref{blackbox-description}. When sampling a problem, and then a hyperparameter configuration at random, we hit a feasible point with probability 51.5\%. Also note that for all these problems, the overall global minimum point is unfeasible.

\begin{table}[]
\centering
\begin{tabular}{llcccc}
Model                 & Dataset    & Constraint       & Threshold   & Feasible points & $d$ \\
\hline
XGBoost               & mg         & model size       & 50000 bytes & 72\% & 7 (5, 2, 0)      \\
Decision tree         & mpg        & model size       & 3500 bytes  & 48\% & 4  (2, 1, 1)         \\
Random forest         & pyrim      & model size       & 5000 bytes  & 26\% & 4 (1, 2, 1)        \\
Random forest         & cpusmall   & model size       & 27000 bytes & 80\% & 4(1, 2, 1)           \\
MLP                   & pyrim      & model size       & 27000 bytes & 79\% & 11 (5, 5, 1)         \\
kNN + rnd. projection             & australian & model size       & 28000 bytes & 29\% & 5 (1, 1, 3)          \\
\hline
MLP                   & heart      & error on neg.\ & 13.3\%      & 30\% & 12  (6, 5, 1)          \\
MLP                   & higgs      & error on neg.\ & 60\%        & 38\% & 12  (6, 5, 1)         \\
Factorization machine & heart      & error on neg.\ & 17\%        & 39\% & 7 (3, 3, 1)         \\
MLP                   & diabetes   & error on neg.\ & 80\%        & 74\% & 12  (6, 5, 1)        \\
\hline
\end{tabular}
\vspace{0.5cm}
\caption{\label{tab:blackbox-description}
   Constrained HPO problems in our experiments. Here $d$ is the input dimension description of the blackbox function, in the following format: total number of dimensions (number of real dimensions, number of integer valued dimensions, number of categorical dimensions). }\end{table}
 
\begin{figure}[]
\vskip -0.075in
\begin{center}
\centerline{
\includegraphics[width=\textwidth]{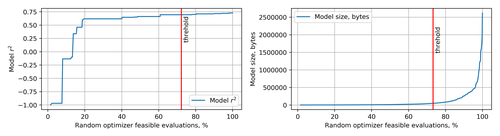}
}
\caption{A tradeoff between model performance ($r^2$) and threshold value for \texttt{XGBoost}, on \texttt{mg} dataset. Smaller threshold results in a smaller number of weak learners, thus degrading the performance of the model.}
\label{figure-threshold-effect}
\end{center}
\vskip -0.25in
\end{figure}

All our example functions require a threshold, either on the size of trained model, or on the error on negatives. For a given HPO problem, the threshold is chosen as follows. First, we sample 2000 random values of the criterion function, without constraint. This allows us to access an effect of a particular threshold on the value of objective, and on a fraction of points which are unfeasible. We select a threshold at random, such that a total fraction of unfeasible points is between 20\% and 80\%. An example visualization for \texttt{XGBoost} is given in Figure \ref{figure-threshold-effect}.

\begin{figure}[h]
\begin{subfigure}{.45\textwidth}
\includegraphics[width=1\textwidth]{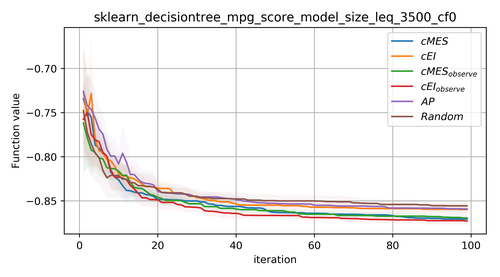}
\end{subfigure}
\begin{subfigure}{.45\textwidth}
\includegraphics[width=1\textwidth]{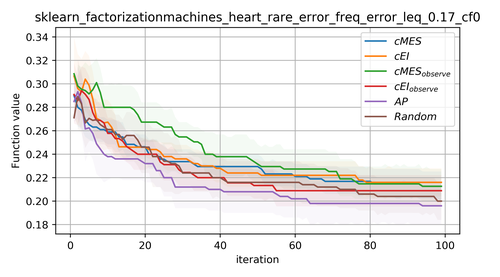}
\end{subfigure}

\begin{subfigure}{.45\textwidth}
\includegraphics[width=1\textwidth]{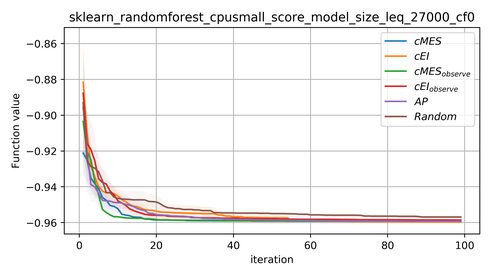}
\end{subfigure}
\begin{subfigure}{.45\textwidth}
\includegraphics[width=1\textwidth]{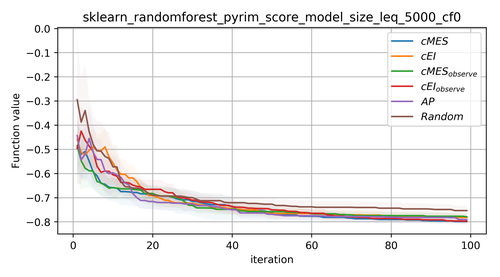}
\end{subfigure}

\begin{subfigure}{.45\textwidth}
\includegraphics[width=1\textwidth]{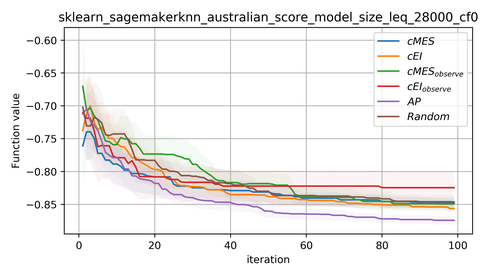}
\end{subfigure}
\begin{subfigure}{.45\textwidth}
\includegraphics[width=1\textwidth]{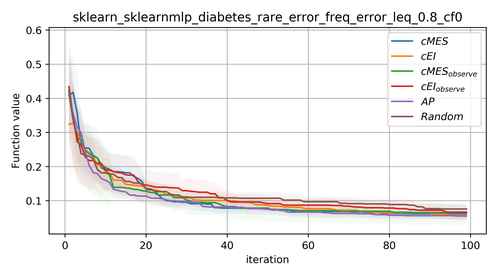}
\end{subfigure}

\begin{subfigure}{.45\textwidth}
\includegraphics[width=1\textwidth]{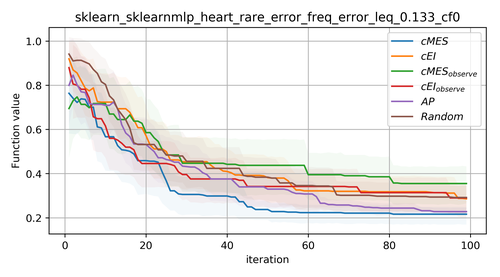}
\end{subfigure}
\begin{subfigure}{.45\textwidth}
\includegraphics[width=1\textwidth]{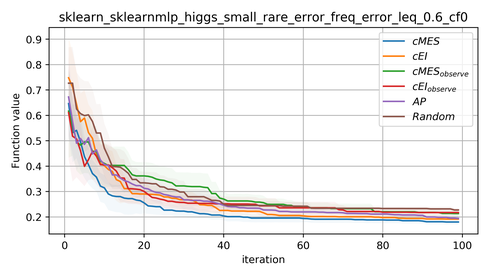}
\end{subfigure}

\begin{subfigure}{.45\textwidth}
\includegraphics[width=1\textwidth]{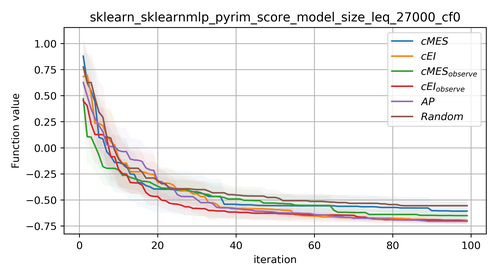}
\end{subfigure}
\begin{subfigure}{.45\textwidth}
\includegraphics[width=1\textwidth]{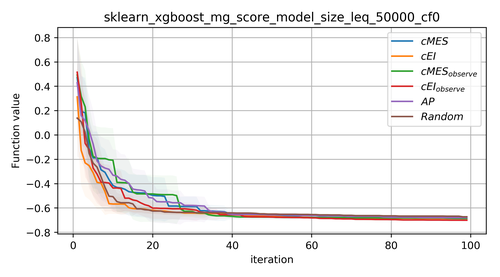}
\end{subfigure}
\caption{Current optimum per iteration for the best-performing methods in each
   category.}
\label{fig:individual_results}
\end{figure}

The results for binary feedback on each individual blackbox are given in Figure~\ref{fig:individual_results}, where we report the current minimum found up to each BO iteration for each competing method. Results are averaged over 20 repetitions and 95\% confidence intervals on the minimal objective value are computed via boostrap.\footnote{www.github.com/facebookincubator/bootstrapped.}

\end{document}